\documentclass[lettersize,journal]{IEEEtran}
\usepackage{amsmath,amsfonts}
\usepackage{algorithmic}
\usepackage{array}
\usepackage[caption=false,font=normalsize,labelfont=sf,textfont=sf]{subfig}
\usepackage{textcomp}
\usepackage{stfloats}
\usepackage{url}
\usepackage{verbatim}
\usepackage{xcolor}
\usepackage{graphicx}

\usepackage{breakurl}
\usepackage[breaklinks]{hyperref}
\usepackage[numbers]{natbib}
\def\BibTeX{{\rm B\kern-.05em{\sc i\kern-.025em b}\kern-.08em
    T\kern-.1667em\lower.7ex\hbox{E}\kern-.125emX}}
\usepackage{balance}
\newcommand{\RN}[1]{%
  \textup{\uppercase\expandafter{\romannumeral#1}}%
}
\begin{document}
\title{A Digital Twin Framework for Physical-Virtual Integration in V2X-Enabled Connected Vehicle Corridors}

\author{Keshu Wu*, Pei Li*, Yang Cheng,  Steven T. Parker, Bin Ran, David A. Noyce, and Xinyue Ye
\thanks{*K. Wu and P. Li contributed equally.}
\thanks{K. Wu was previously with the Department of Civil and Environmental Engineering at the University of Wisconsin-Madison, Madison, WI, 53706, USA. He is currently affiliated with the Center for Geospatial Sciences, Applications, and Technology, as well as the Department of Landscape Architecture and Urban Planning, and the Zachry Department of Civil and Environmental Engineering, Texas A\&M University, College Station, TX, 77840, USA (email: keshuw@tamu.edu).}
\thanks{P. Li, C. Yang, S. Parker, B. Ran, and D. Noyce are all associated with the Department of Civil and Environmental Engineering at the University of Wisconsin-Madison, Madison, WI, 53706, USA (e-mail: pei.li@wisc.edu, cheng8@wisc.edu,  sparker@engr.wisc.edu, bran@wisc.edu, and danoyce@wisc.edu)}
\thanks{X. Ye is affiliated with the Center for Geospatial Sciences, Applications, and Technology, the Department of Landscape Architecture and Urban Planning, as well as the Department of Computer Science and Engineering, Texas A\&M University, College Station, TX, 77840, USA (e-mail: xinyue.ye@tamu.edu).}
}


\maketitle

\begin{abstract}
Transportation Cyber-Physical Systems (T-CPS) enhance safety and mobility by integrating cyber and physical transportation systems. A key component of T-CPS is the Digital Twin (DT), a virtual representation that enables simulation, analysis, and optimization through real-time data exchange and communication. Although existing studies have explored DTs for vehicles, communications, pedestrians, and traffic, real-world validations and implementations of DTs that encompass infrastructure, vehicles, signals, communications, and more remain limited due to several challenges. These include accessing real-world connected infrastructure, integrating heterogeneous, multi-sourced data, ensuring real-time data processing, and synchronizing the digital and physical systems. To address these challenges, this study develops a traffic DT based on a real-world connected vehicle corridor. Leveraging the Cellular Vehicle-to-Everything (C-V2X) infrastructure in the corridor, along with communication, computing, and simulation technologies, the proposed DT accurately replicates physical vehicle behaviors, signal timing, communications, and traffic patterns within the virtual environment. Building upon the previous data pipeline, the digital system ensures robust synchronization with the physical environment. Moreover, the DT’s scalable and redundant architecture enhances data integrity, making it capable of supporting future large-scale C-V2X deployments. Furthermore, its ability to provide feedback to the physical system is demonstrated through applications such as signal timing adjustments, vehicle advisory messages, and incident notifications. The proposed DT is a vital tool in T-CPS, enabling real-time traffic monitoring, prediction, and optimization to enhance the reliability and safety of transportation systems.
\end{abstract}


\section{Introduction}
\IEEEPARstart{T}{ransportation} Cyber-Physical Systems (T-CPS) have emerged as a transformative force in enhancing safety and mobility by enabling feedback-based interactions between the cyber and physical systems~\cite{xiong2015cyber, rawat2015towards}. Leveraging sensing, computing, simulation, and communication technologies, T-CPS links the physical infrastructure with the virtual systems, allowing seamless communication and interactions between them. In particular, Vehicle-to-Everything (V2X) is one of the crucial technologies in T-CPS. V2X enables vehicles to communicate with each other, pedestrians, cyclists, and roadside infrastructure through wirelessly exchanged messages. V2X provides road users with real-time information about traffic situations, allowing responses that could improve safety, optimize system performance, and enhance mobility~\cite{v2x_usdot}.

\begin{figure}[!t]
    \centering
    \includegraphics[width=0.5\textwidth]{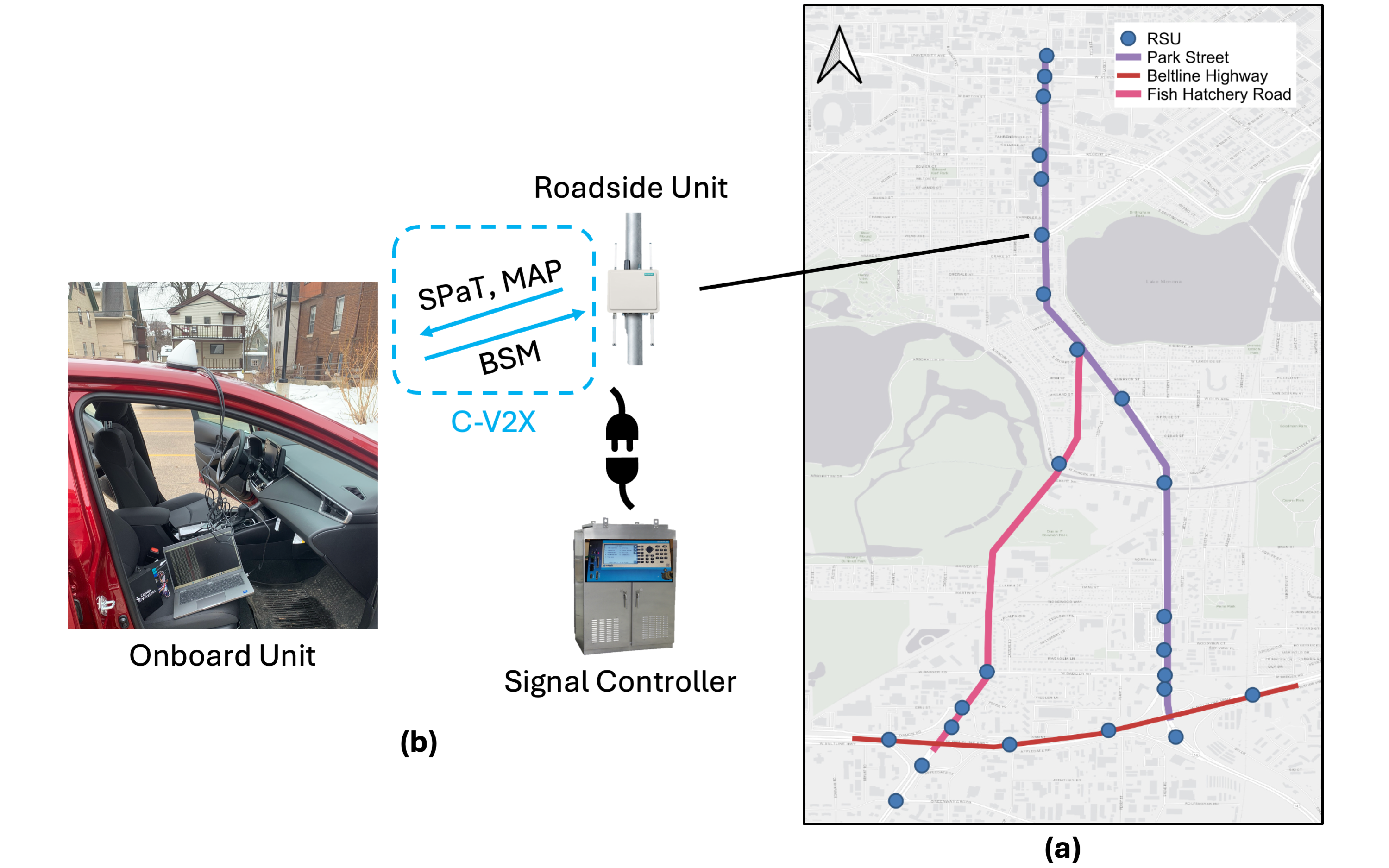}
    \caption{CV Corridor Deployment in Madison, WI, USA. (a) The layout of the CV corridor, (b) The data transmission among the RSU, signal controller, and OBU.}
    \label{fig_deploy}
\end{figure}

In recent years, V2X has been deployed across the globe, creating real-world test beds to develop and validate various T-CPS applications. Fig.~\ref{fig_deploy}(a) presents the Park Street Connected Vehicles (CV) corridor used in this study. Park Street spans from University Avenue to the US-12/18 Freeway in Madison, serving as a critical corridor connecting the freeway to downtown Madison, the UW campus, and several regional medical facilities. This heavily utilized corridor, which accommodates approximately 48,000 vehicles daily, is an ideal environment for piloting and studying V2X applications~\cite{park_corridor}.

The Park Street CV corridor was initiated in response to the SPaT Challenge organized by AASHTO, ITE, and ITS America in 2017. This challenge aimed to deploy Dedicated Short-Range Communications (DSRC) in every US state~\cite{spat_challenge}. The corridor was first equipped with Roadside Units (RSUs) with DSRC capabilities. As shown in Fig.~\ref{fig_deploy}(b), RSUs are vital units that enable the communication between infrastructure and vehicles. Each RSU is wired to a local signal controller and broadcasts V2X messages to other RSUs and Onboard Units (OBUs). The number of RSUs was selected based on the corridor’s signalized intersections and its geographic and traffic characteristics, ensuring sufficient coverage for communication. Subsequently, Cellular Vehicle-to-Everything (C-V2X) emerged as an alternative to DSRC, offering enhanced capabilities. Unlike DSRC, which supports only direct communication, C-V2X supports both short-range, direct and long-range, indirect communication. This dual-mode approach enables more flexible communication, better coverage in rural areas, and reduced reliance on additional RSU deployments. In 2022, the RSUs on Park Street were reconfigured to incorporate C-V2X capabilities, aligning with the transition in V2X communication preferences among policymakers, researchers, and original equipment manufacturers~\cite{fcc_2024}. 

\begin{figure*}[!t]
\centering
\includegraphics[width=0.8\textwidth]{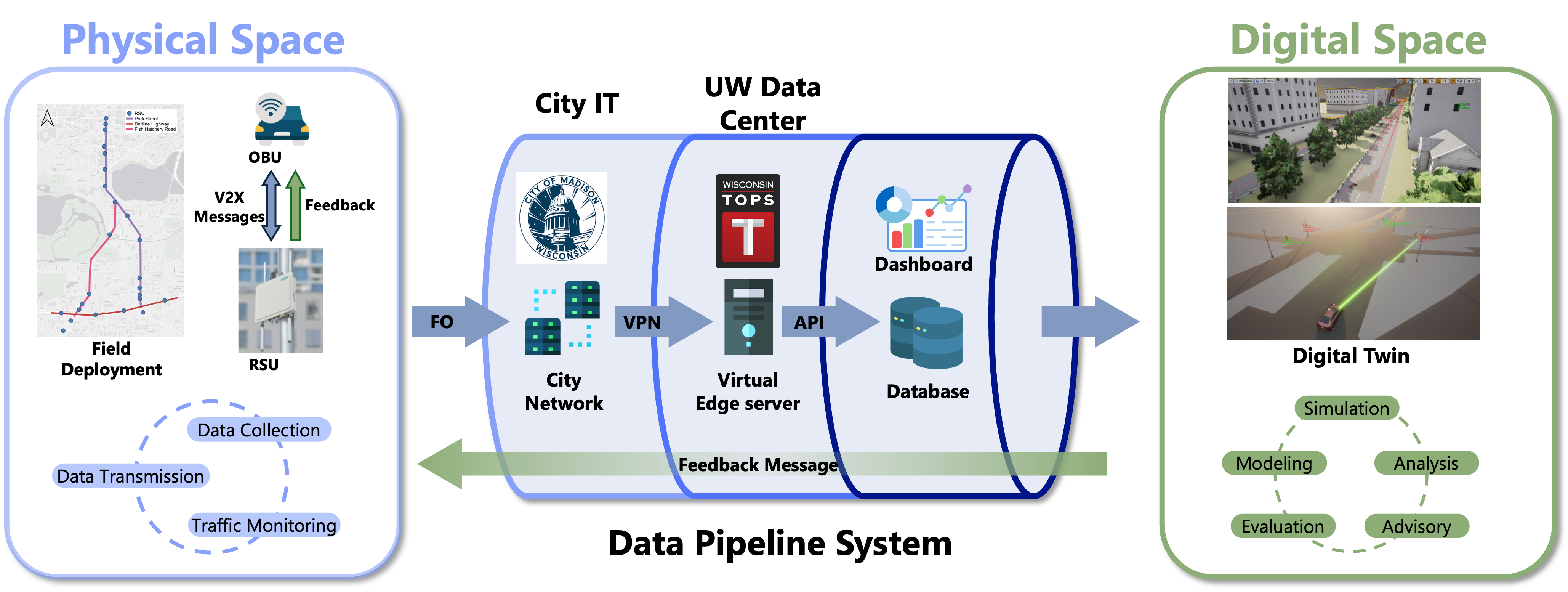}
\caption{Connected Vehicle Corridor Data Pipeline and Digital Twin.}
\label{fig_datapipeline}
\end{figure*}

The CV corridor provides essential V2X infrastructure, but processing the vast amounts of data generated by infrastructure, vehicles, and road users remains a challenge. To address this, the authors have developed a data pipeline as shown in Fig.~\ref{fig_datapipeline}. This pipeline utilizes the University of Wisconsin-Madison’s computing infrastructure to manage data~\cite{wu2023development, li2023does}. A dedicated virtual private network ensures uninterrupted network connectivity between the pipeline and the RSUs, forming a seamless data acquisition and archival system. 

Leveraging the existing data pipeline, this study develops a traffic Digital Twin (DT) to advance future T-CPS applications. A DT is a virtual representation of a physical system that facilitates simulation, analysis, prediction, and optimization. It comprises three main components: (1) A physical reality, (2) a virtual representation, and (3) interconnections that exchange information between the physical reality and virtual representation \cite{VANDERHORN2021113524, kritzinger2018digital, jones2020characterising, fuller2020digital}. DTs have been developed across diverse domains from manufacturing to healthcare and disaster response, with the transportation sector becoming an increasing area of their applications \cite{Almeaibed2021, Wang2022, Schwarz2022, hui2022collaboration, niaz2021autonomous, hu2022review, dong2023mixed,  ye2023developing, cai2023, WAGNER2023, liao2024digital, wang2024smart, cazzella2024multi}.

Existing studies have focused on developing DTs for vehicle control, human behavior modeling, and communication optimization, while research on traffic DTs remains limited. A traffic DT replicates the transportation ecosystem in the digital space, enabling proactive issue identification, solution validation, and operational optimization. Its feedback loop supports continuous system improvements, making it a crucial tool in T-CPS. Moreover, most existing studies have developed and evaluated transportation DTs in simulated environments. However, developing traffic DTs that integrate infrastructure, signal, and vehicle data in real-world environments faces additional challenges. First, handling the immense volume of data generated by infrastructure, vehicles, and road users requires advanced retrieval, processing, and storage solutions. Second, the real-time operation of the DT demands instantaneous processing of V2X messages, as well as advanced techniques for data synchronization and validation to ensure timely and accurate responses. Third, the diverse nature of the data, ranging from structured packets to unstructured sensor readings, poses significant integration challenges, requiring robust techniques for data integration.

To address these challenges, we first enhance the existing data pipeline to integrate and synchronize various real-time V2X messages, including Signal Phase and Timing (SPaT), Intersection Mapping (MAP), and Basic Safety Messages (BSM). A traffic DT is developed using CARLA (Car Learning to Act) \cite{dosovitskiy2017carla}, an open-source platform widely used for autonomous driving research. As shown in Fig.~\ref{fig_datapipeline}, the DT replicates the physical transportation system, creating a virtual environment with various entities such as vehicles, intersections, roadways, and traffic signals. By assimilating real-time V2X messages, the DT replicates traffic conditions, analyzes traffic data, and provides advisory information to the physical space. Our main contributions are summarized as follows:

\begin{itemize}
    \item We present the architecture and key components of the enhanced data pipeline. The pipeline integrates V2X data using advanced temporal synchronization and spatial matching techniques.
    \item We present a traffic DT developed based on the real-world corridor. The DT maps physical data into a virtual environment, ensuring that virtual entities such as vehicles, OBUs, RSUs, traffic signals, and V2X communications accurately replicate their real-world counterparts.
    \item We evaluate the data pipeline and DT through extensive experiments, showcasing improvements in data integration, data synchronization, and simulation fidelity.
    \item We demonstrate the DT's ability to generate actionable feedback for the physical environment, encompassing signal timing adjustments, vehicle advisory messages, and incident notifications.
\end{itemize}


The remainder of the paper is organized as follows: Section \RN{2} reviews existing research on DTs in transportation. Section \RN{3} details the architecture and essential components of the enhanced data pipeline and DT. Section \RN{4} provides evaluation results of the DT, focusing on data integration, replication, and synchronization. Section \RN{5} highlights system features and discusses potential improvements. Section \RN{6} presents the conclusions of the paper.

\section{Related Work}

Existing studies in transportation DTs have four main focus areas, including vehicle~\cite{wang2021digital,hui2022collaboration,dong2023mixed,tan2023digital,liao2024digital,wang2024smart}, communication~\cite{cai2023,zelenbaba2022wireless, liu2022application, tan2023digital, cai2023, cazzella2024multi}, traffic~\cite{WAGNER2023, dasgupta_transportation_2023, adarbah_digital_2024, adarbah_digital_2023}, and pedestrian~\cite{wang2023towards, fu2024digital} DTs.

Most studies in vehicle DTs have focused on cooperative driving automated (CDA) and advanced driver assistance systems (ADAS). For instance, \citet{wang2021digital} developed a DT that assimilates onboard sensor data and predicts surrounding vehicle behavior to assist ego-vehicle's decision-making.~\citet{hui2022collaboration} developed a DT system to optimize CDA by minimizing costs and reducing information exchanges between multiple AVs in a coalition. The authors created DTs of individual AVs. Two optimization mechanisms were developed to decide each coalition's head and tail DT and determine the optimal coalition distribution, helping AVs make collaborative driving decisions. \citet{dong2023mixed} developed a DT for validating and testing multi-vehicle CDA. The physical space was created using a sand table that contains miniature vehicles, roadside sensors, and other facilities. Digital replicas of the physical entities were created using the Unity engine. The authors have validated the proposed system's feasibility through multi-vehicle platooning experiments. \citet{liao2024digital} proposed a DT to facilitate CDA by enabling macroscopic traffic optimization. The DT collects real-time traffic conditions and provides global-optimized routes for AVs. The global policy iteration adapts the AV cooperation mechanisms to macroscopic traffic flow and density, maximizing global benefits. Similarly,~\citet{wang2024smart} proposed a DT-based navigation system for CAVs. The DT collects real-time traffic data from RSUs and CAVs, providing route suggestions to CAVs to avoid crashes and other incidents. The authors have validated the proposed DT using three RSUs and three CAVs within a campus environment. Moreover, some studies have explored other applications of vehicle DTs. \cite{Almeaibed2021} developed a theoretical DT framework that detects cyber-attacks on AVs by assimilating data from AV sensors and performing safety analysis.

Efficiency and resource allocations are the two main focuses in communication DTs. For example, \citet{zelenbaba2022wireless} developed a DT to emulate V2V communication. Using metrics including latency and packet error rate, the DT evaluates the performance of vehicular communication. The authors have used the DT to select the appropriate communication techniques for transmitting safety-critical information. \citet{liu2022application} proposed a conceptual framework for a lightweight DT to enhance its efficiency and reduce the cost of collecting unnecessary data. Several strategies, including multi-sensor fusion, real-time data abstraction, and dynamic communication resource allocation were applied to enhance communication efficiency. Similarly, \citet{tan2023digital} proposed a DT architecture for vehicle-to-cloud communication. By integrating and reusing resources within vehicular networks, the DT optimizes V2X communication, system reliability, and resource allocation. Moreover, \citet{cai2023} proposed a DT framework to perform efficiency-oriented V2X network scheduling. They demonstrated the DT's role in driver behavior-based V2X scheduling and cooperative vehicle merging, achieving improved task efficiency with reduced complexity. For example, by identifying driver behavior, the framework optimizes communication resource allocation, prioritizing aggressive drivers for timely safety warnings via V2X. Lastly, \citet{cazzella2024multi} developed a DT framework to improve existing V2X communications by allowing the control of physical equipment based on accurate virtual sensors and channel data. The authors have validated the DT's effectiveness in aiding blockage handover.

Some studies have developed traffic DTs with a focus on traffic signal control.~\citet{WAGNER2023} developed a DT using SUMO and OMNeT++. The authors simulated V2I communication with a focus on SPaT and MAP messages. Moreover, artificial signal plans were used in this study as the authors did not have access to real-world data. \citet{dasgupta_transportation_2023} developed a DT-based traffic signal control system. The DT replicates city-wide traffic signals, offering a global perspective for optimizing signal control. The authors validated the control algorithm using a simulated environment developed in SUMO. Similarly,~\citet{adarbah_digital_2024} proposed a DT-based signal control system to monitor traffic conditions including flow and vehicle behavior. The system reduced fuel consumption and stop time compared to existing studies. In addition, \citet{adarbah_digital_2023} have discussed the potential of 5G in traffic DTs. A conceptual architecture of DT was developed while considering scalability, reliability, and interoperability.

Lastly, a few studies have developed DTs for pedestrians. \cite{wang2023towards} developed a DT framework for connected vehicles and pedestrian in-the-loop simulation. The digital space is created based on historical data collected using drones. Moreover, simulators are used to capture the behavior of drivers and pedestrians, creating digital replicas in the digital space. \citet{fu2024digital} developed a DT for protecting pedestrians at intersections. Cameras and LiDARs were used to extract real-time vehicle and pedestrian behavior for estimating surrogate safety measures. Road users were replicated in a simulated environment in CARLA to validate the proposed warning strategies.

Table \ref{table:relevant_studies} has summarized existing studies from various aspects. In summary, most literature focuses on cooperative vehicle control, interactions between vehicles and other road users, and communications, while some studies have developed traffic DTs with a focus on signal control. However, several key gaps still require further research. First, most studies validate DTs in simulation without real-world implementation, which limits their applicability to dynamic and unpredictable traffic conditions. Second, while DTs for individual vehicles and localized intersections exist, city-wide or corridor-scale DTs that integrate real-time signal status, vehicle behavior, and infrastructure conditions are limited. Developing traffic DTs that interact with physical transportation systems presents additional challenges, including data acquisition, real-time synchronization, sensor fusion, scalability, and interoperability with existing infrastructure systems.


\begin{table*}[ht]
\caption{Existing Studies on Transportation DTs} 
\centering
\begin{tabular}{llllll}
\hline
{Focus Area} & {Study} & {Application} & {Implementation} \\
\hline
    Vehicle
            & \citet{wang2021digital} & Behavior prediction for ADAS & Simulation \\
            & \citet{Almeaibed2021} & AV cyber-attack detection & Simulation \\
            & \citet{hui2022collaboration} & Collaborative and distributed autonomous driving & Simulation\\
            & \citet{dong2023mixed} & Mixed cloud control testbed for CDA & Sand table test bed \\
            & \citet{cai2023} & Driver monitoring and cooperative merging & Simulation\\
            & \citet{wang2024smart} & CAV navigation system & Real-world campus environment \\
            & \citet{liao2024digital} & Edge-to-Cloud architecture for traffic guidance & Simulation \\
    Communication 
            & \citet{zelenbaba2022wireless} & Vehicular communication emulation & Simulation\\
            & \citet{liu2022application} & Communication resource optimization & Concept \\
            & \citet{tan2023digital} & Resource integration in vehicular networks & Simulation\\
            & \citet{cazzella2024multi} & Blockage handover for V2X link restoration & Simulation\\
    Traffic 
            & \citet{WAGNER2023} & DT for simulating SPaT and MAP messages & Simulation\\
            & \citet{dasgupta_transportation_2023} & DT for signal control & Simulation\\
            & \citet{adarbah_digital_2023} & Traffic DT architecture with 5G-V2X & Concept\\
            & \citet{adarbah_digital_2024} & DT for signal control & Simulation\\
    Pedestrian 
            & \citet{wang2023towards} & DT framework for studying vehicle-pedestrian interactions & Simulation \\
            & \citet{fu2024digital} & Pedestrian warning validation & Simulation \\

\hline
\label{table:relevant_studies}
\end{tabular}
\end{table*}

\section{Enhanced Data Pipeline for Digital Twin} 

The previous data pipeline~\cite{wu2023development} has laid the groundwork for developing the DT. This section enhances the pipeline, facilitating real-time simulations and in-depth analysis of complex interactions between infrastructure, vehicles, and road users. 

\subsection{V2X Message Types} 
The Society of Automotive Engineers (SAE) defines the J2735-2016 standard \cite{sae-j2735_2016}, which governs the messages, data frames, and data elements for V2X. This study focuses on three message types that are essential for developing the DT:

\begin{enumerate} 
    \item \textbf{SPaT}: This dynamic message communicates the current and future status of signalized intersections. It broadcasts at 10 Hz, providing real-time information essential for safety-critical decision-making. 
    \item \textbf{MAP}: This static message describes the geometric layout of an intersection, including lane-level details. Broadcast at 1 Hz, the MAP message complements SPaT by providing accurate road layout and lane alignment. 
    \item \textbf{BSM}: This dynamic message contains a vehicle’s status, such as its location, speed, and heading, and is broadcast at 10 Hz. 
\end{enumerate}

\subsection{Data Pipeline Architecture} 
Fig. \ref{fig_arc}(a) presents the first version of the data pipeline \cite{wu2023development}. This paper enhances the data pipeline for developing the DT, as shown in Fig.~\ref{fig_arc}(b). Key upgrades include temporal synchronization and spatial matching of V2X messages. The enhanced system integrates these features through six core modules designed to ensure safety and reliability in the data pipeline for T-CPS:

\begin{figure*}[!ht]
  \centering
  \includegraphics[width=0.8\textwidth]{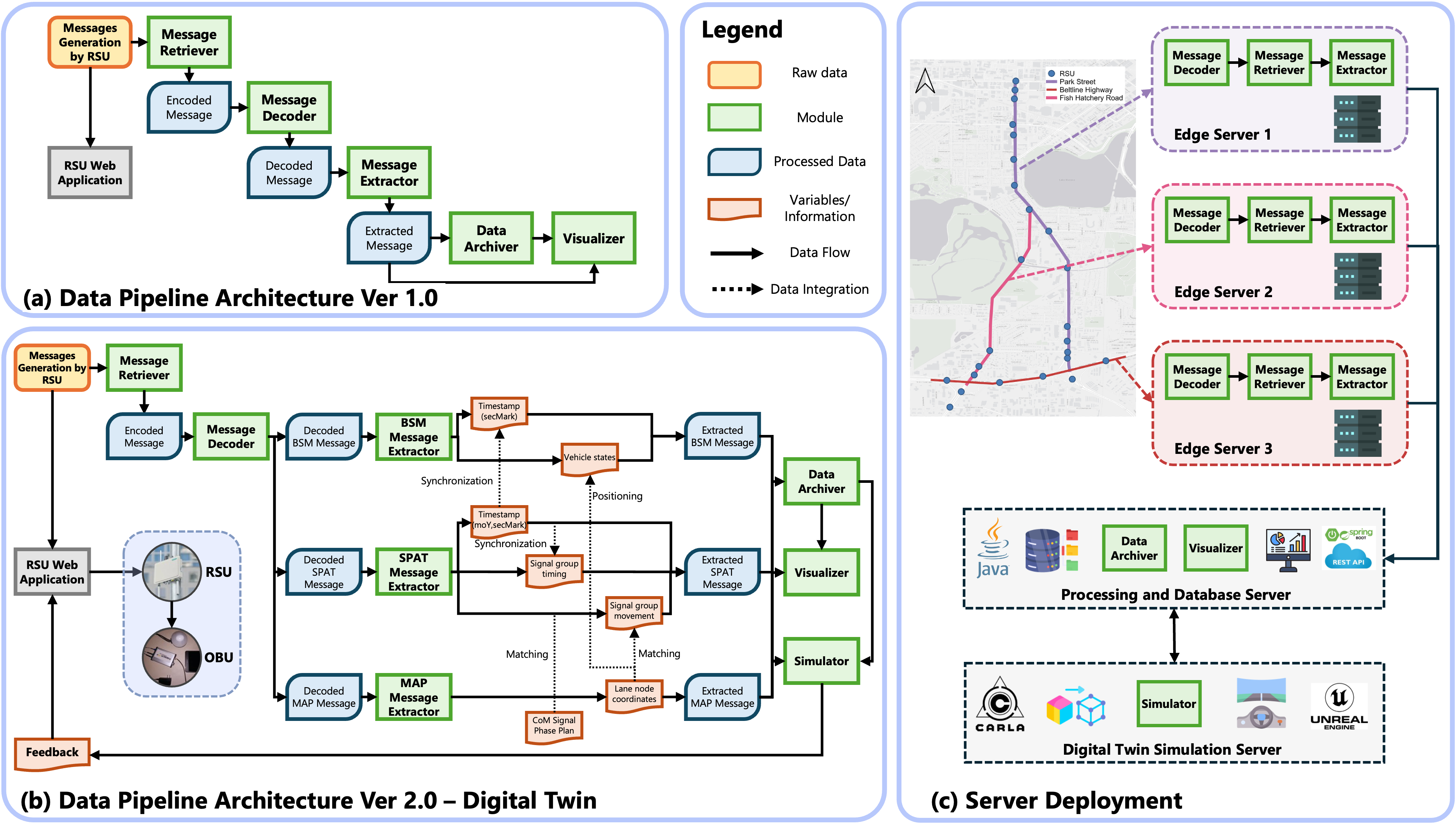}
  \caption{Data Pipeline Architecture}
  \label{fig_arc}
\end{figure*}

\begin{enumerate} 
    \item \textbf{Message Retriever}: This module continuously communicates with RSUs to collect real-time V2X messages. It manages message reception via JSON commands and processes messages encoded using the Unaligned Packed Encoding Rules (UPER). 
    \item \textbf{Message Decoder}: The decoder translates payload strings into readable data structures. By utilizing OSS ASN.1 Tools for Java, this module decodes hex strings into JSON objects, making V2X messages understandable and generating separate JSON files for each message. 
    \item \textbf{Message Extractor}: These extractors process JSON strings to convert them into a tabular format for ease of analysis. The extractors also fuse data elements from V2X messages, organizing critical information into a new JSON object that simplifies further processing. In this module, the adoption of dedicated extractors for each message type enables immediate synchronization and fusion during extraction, improving efficiency, ensuring alignment at the source, and enhancing accuracy for downstream tasks.
    \item \textbf{Data Archiver}: The archiver manages short- and long-term data storage using a PostgreSQL database. It saves decoded records for short-term analysis while filtering and archiving only relevant data for long-term storage. The archiver’s structure ensures that both static and dynamic data are maintained for future retrieval.
    \item \textbf{Visualizer}: The visualizer, developed using HTML and JavaScript, displays real-time and historical SPaT data on a dashboard. It functions via the Spring Boot framework for web services, ensuring seamless access to live and archived data. 
    \item \textbf{Simulator}: The simulator transfers data from the physical environment into the virtual simulation. The CARLA simulator is adopted to create the simulation environment. The states of the virtual OBUs, RSUs, and traffic signals replicate real-world configurations, maintaining a high degree of fidelity. 
\end{enumerate}

Moreover, Fig.~\ref{fig_arc}(c) illustrates the deployment of the three edge servers, physically located at the University of Wisconsin-Madison campus. Each edge server is responsible for managing a set of RSUs. Edge Server 1 processes data from RSUs on Park Street, Edge Server 2 handles RSUs along Fish Hatchery Road, and Edge Server 3 manages RSUs on the Beltline Freeway. Each edge server hosts modules for message reception, decoding, and extraction, ensuring efficient data processing. Once the RSUs establish connections with their designated edge servers, the processed data is transmitted to the processing and database server. This server houses the Data Archiver and Visualizer, where data is stored and analyzed. The Data Archiver securely logs information in the database, while an API enables seamless data retrieval and visualization, maintaining connectivity between the Visualizer and Archiver. Additionally, the DT is deployed on the Digital Twin Simulation Server. This system integrates CARLA, Unreal Engine, and the simulation framework to replicate real-world traffic conditions. The simulation environment interacts with the database and processing server, allowing real-time data synchronization and validation, ultimately enhancing traffic analysis and decision-making. Moreover, as shown in Fig.~\ref{fig_arc}(b), our system leverages a web application for RSU devices, which allows for uploading and broadcasting messages directly from RSUs to OBUs in real-time. This web application ensures seamless communication of feedback from the virtual environment to the physical space.

\subsection{Data Extractor Enhancement} 
\begin{figure}[!ht]
  \centering
  \includegraphics[width=0.5\textwidth]{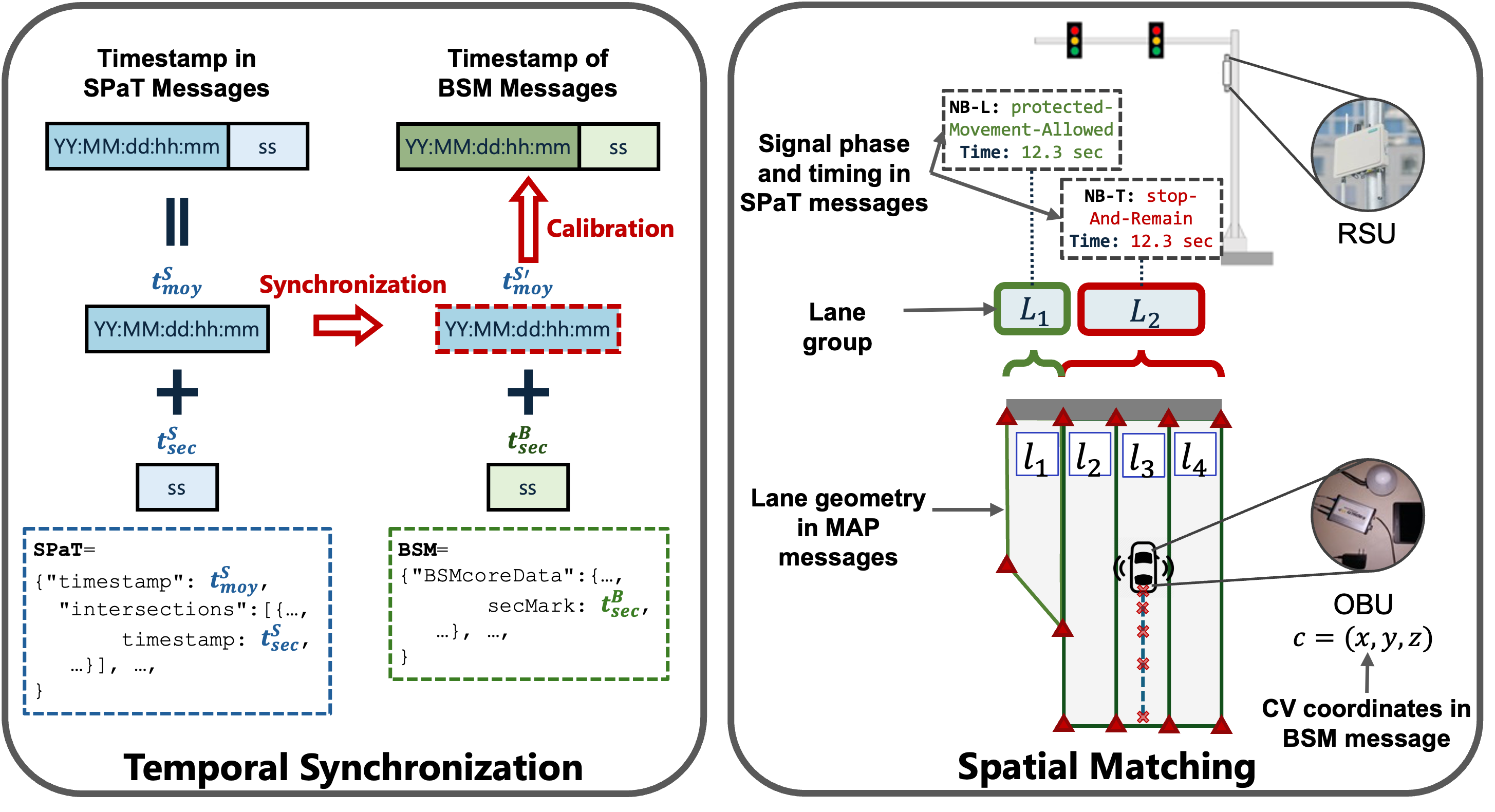}
  \caption{Data Extractor Module: Temporal Synchronization and Spatial Matching}
  \label{fig_data_extract}
\end{figure}

The previous data extractor focused on extracting elements from structured data frames, without integrating the additional information embedded within the message content~\cite{wu2023development}. In this work, we enhance the message extractor module by adding two key functionalities: temporal synchronization and spatial matching, as illustrated in Fig.~\ref{fig_data_extract}. These enhancements integrate V2X messages both temporally and spatially, enabling the module to generate a more comprehensive and accurate dataset. This improvement supports more reliable simulations by ensuring the data is aligned across time and space.

\begin{table}[!t]
        \scriptsize
	\caption{Signal Group and Movement of RSU at two Intersections}\label{tab_phase}
	\begin{center}
		\begin{tabular}{l l l l l}
            \hline 
			Intersection & Signal Group & Movement & Signal Group & Movement\\\hline
		      \#1 & 1 & SB-L & 5 & NB-L\\
                      & 2 & NB-T & 6 & SB-T\\
                      & 3 & EB-L & 7 & WB-L\\
                      & 4 & WB-T & 8 & EB-T\\\hline
                \#2 & 2 & NB-T & 6 & SB-T\\
                      & 3 & SB-L & 9 & WB-R\\
                      & 4 & WB-T &  & \\\hline
		\end{tabular}
	\end{center}
\end{table}

\begin{table}[!ht]
        \scriptsize
	\caption{Integrated V2X Message Data Elements}\label{tab_messages}
	\begin{center}
		\begin{tabular}{p{0.5cm} p{1.2cm} p{0.4cm} p{5.0cm}}
  \hline
			Message & Variable & Type & Description \\\hline
			BSM & timestamp & date & Message generated time\\
                & latitude & double & The latitude of the vehicle in degrees\\
                & longitude & double & The longitude of the vehicle in degrees\\
                & elevation & double & The elevation of the vehicle in cm\\
                & speed & double & The vehicle speed in m/s\\
                & heading & double & The heading direction of motion in degrees\\
                & width, length & integer & The vehicle width and length in cm\\
                & accuracy & integer & The accuracy of positional determination\\
                & acceleration & integer & The acceleration status in three orthogonal directions\\
                & angle & double & The driver's steering angle in degrees \\
                & brakeStatus & string & The brake and system control activity\\
                & transmission & string & The current state of transmission\\
                & pathHist & double & List of historical path points\\
                & pathPred & double & Predicted path point\\
                & intersectionID & integer & The ID of the intersection the vehicle located \\
                & signalGroupID & integer & The signal group ID governing the lane(s) where the vehicle is located\\\hline
			SPaT & intersectionID & integer & The ID of the intersection  \\
                & signalGroupID & integer & The signal group ID governing one or more lanes at the intersection\\
			& timeStamp & date & Message generated time\\
                & eventState & string & The phase state of the signal\\
                & minEndTime & double & Expected shortest end time in seconds \\
                & maxEndTime & double & Expected longest end time in seconds \\\hline
			MAP & refPoint & double & The coordinate of the reference point\\
                & laneWidth & double & The width of a lane\\
			& nodeList & double & List of node points in lane spatial path\\
			& signalGroupID & integer &  The signal group ID governing the associated lane(s)\\
                & connectingLane & integer & Lane index of the connecting lane\\\hline
		\end{tabular}
	\end{center}
\end{table}

\subsubsection{Temporal Synchronization}  
As illustrated in Fig.~\ref{fig_data_extract}, temporal synchronization aligns SPaT and BSM messages based on their respective timestamps. SPaT messages include three key temporal elements: (1) $t_{moy}^S$, representing the number of minutes elapsed in the current year when the message was generated according to the system clock, (2) $t_{sec}^S$, indicating the number of milliseconds within the current UTC minute at the time of message creation, and (3) $t_{end}^S$, which specifies the estimated time—expressed in tenths of a second within the current hour—when the signal phase for a given signal group is expected to change. The superscripts indicate that these parameters are derived from SPaT messages. To process this temporal information, $t_{moy}^S$ is first converted into milliseconds and then combined with $t_{sec}^S$ to determine the absolute timestamp of the SPaT message:
\begin{equation}
    t^S = t_{moy}^S + t_{sec}^S
\end{equation}  

This computed value is then formatted into a standard date-time representation. Additionally, to determine the remaining duration until the signal phase change, $t_{end}^S$ is converted into milliseconds, and the residual time is computed as:  
\begin{equation}
    t_{end;p}^S = t_{end}^S - (t_{moy}^S + t_{sec}^S)
\end{equation}

Where $t_{end;p}^S$ represents the remaining time (in milliseconds) until the signal phase change occurs, while the sum $t_{moy}^S + t_{sec}^S$ denotes the precise timestamp at which the signal phase change is expected to occur.

Conversely, BSM messages include a single time descriptor, $t_{sec}^B$, which records the number of milliseconds elapsed within the current UTC minute when the message is generated. To compute the full timestamp for a BSM message, the most recent SPaT message is referenced to obtain $t_{moy}^{S'} = t_{moy}^S$, and the timestamp is determined as follows:  
\begin{equation}
    t^B = t_{moy}^{S'} + t_{sec}^B
\end{equation}  
where $t^B$ represents the precise generation timestamp of the BSM message. Unlike SPaT messages, MAP messages are updated infrequently and only when there are geometric changes in the road network. As a result, real-time synchronization is not necessary for MAP messages.

\subsubsection{Spatial Matching}
In addition to temporal synchronization, matching V2X messages spatially plays a critical role in correlating real-time traffic data with physical lane geometries and signal states. Fig.~\ref{fig_data_extract} gives an example in which spatial matching is used to achieve lane-level vehicle positioning and signal phase verification. SPaT messages include both temporal information and the present state of the signal groups, represented by the element $s$. The City of Madison controller plan defines the signal group movements for each intersection, as shown in Table \ref{tab_phase}. These movements are combined to create signal phase designs that accommodate various traffic flows. Moreover, MAP messages provide detailed information about lane geometries, signal groups, and lane connections. For a given intersection with $n$ lanes, let $L$ represent the set of lane geometries, where each lane $l_i$ is represented as a polygon with $m$ nodes:
\begin{equation}
    L = \{l_i|\forall i \in[1,\dots,n]\}
\end{equation}
where $l_i$ signifies the polygon of the lane geometry for lane $i$. Assuming the polygon comprises $m$ nodes, we can further designate
\begin{equation}
    l_i = \{c_i^j|\forall j \in[1,\dots,m]\} = \{(x_i^j, y_i^j,z_i^j)|\forall j \in[1,\dots,m]\}
\end{equation}
where $x_i^j, y_i^j,z_i^j$ demarcate the latitude, longitude, and elevation of the $j$th nodes $c_i^j$ in polygon $i$. It is feasible to associate the lane geometry with its real-time signal group and the phase state using the signal group ID.

BSM messages log the real-time position of vehicles using latitude, longitude, and elevation, represented as $c^B = (x^B, y^B, z^B)$. To determine which lane a vehicle is currently occupying, we use a convex hull algorithm \cite{convexhull}, which takes the vehicle’s coordinates and the lane geometry from the MAP as inputs. The convex hull represents the smallest convex polygon that can encompass all points in a given set. If a vehicle is within a lane, its coordinates will fall inside the convex hull of the lane polygon. A binary indicator $\text{ind}_i$ is used to determine whether the vehicle is located in lane $i$:
\begin{equation}
    \text{ind}_i = \begin{cases}
    1, & \text{if } c^B \text{ lies in } l_i\\
    0, & \text{otherwise}
\end{cases}
\end{equation}
Thus, when a connected vehicle enters a lane at an intersection equipped with an RSU, the signal group associated with the vehicle’s current lane can be identified as: 
\begin{equation}
    s^B = \{ s_i | \text{where }\text{ind}_i=1\}
\end{equation}
This allows the system to associate the vehicle’s position with real-time signal data, providing an integrated view of traffic dynamics and enhancing safety monitoring. Moreover, the structure of the integrated messages, along with their data elements, is detailed in Table \ref{tab_messages}, which provides a comprehensive overview of how BSM, SPaT, and MAP messages are fused and processed within the system.

\subsection{Digital Twin System}


\begin{figure}[!ht]
    \centering
    \includegraphics[scale=0.35]{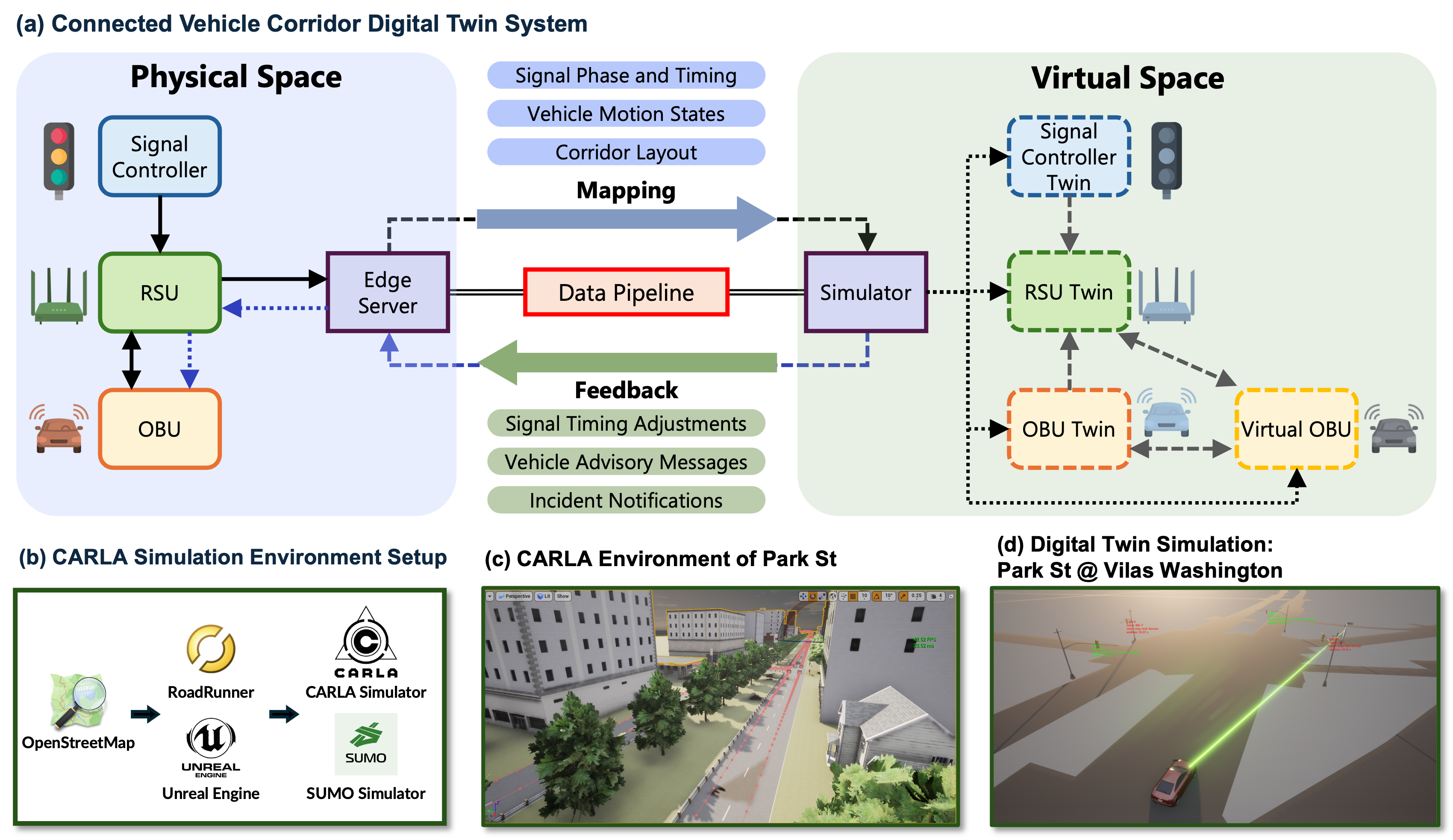}
    \caption{CARLA Simulation and Digital Twin}\label{fig_digitaltwin}
  \end{figure}

\subsubsection{Digital Twin System Design}
This study leverages CARLA to construct the DT. CARLA's flexibility and ability to represent dynamic traffic scenarios make it an ideal choice for replicating physical systems in a virtual environment~\cite{feng2019,hang2021cooperative}. Fig.~\ref{fig_digitaltwin}(a) illustrates the DT's architecture, where real-world data is continuously ingested and synchronized with the virtual environment. The DT is built upon a two-tier system comprising the Physical Space and Virtual Space.

In the Physical Space, traffic signal controllers regulate intersections and communicate SPaT data to RSUs. OBUs transmit BSMs that provide information on vehicle behaviors. These data streams are aggregated at an edge server, which serves as the intermediary between the physical world and the virtual simulation. The edge server processes various V2X messages and forwards them through the data pipeline to ensure seamless integration with the Virtual Space.

In the Virtual Space, the simulator reconstructs a digital representation of the CV corridor by creating twins of RSUs, OBUs, and Signal Controllers. The RSU twins generate and transmit SPaT and MAP messages to the OBU twins, while also receiving other V2X messages. The OBU twins synchronize with live BSM data to reflect the vehicles's position, speed, and behavior. Additionally, virtual OBUs are generated using the CARLA API. Unlike the OBU twins, which strictly mirror real-world data, the Virtual OBUs act as interactive agents that engage dynamically with the OBU twins and virtual traffic signals. They generate synthetic BSM messages, simulating vehicles for the evaluation of additional traffic agents and hypothetical scenarios.

The environment-building process is illustrated in Fig.~\ref{fig_digitaltwin}(b), which transforms OpenStreetMap (OSM) data into a fully operational CARLA simulation. The process begins with extracting road network data from OSM, which is then imported into Unreal Engine for visualization and further refinement using RoadRunner. This refinement ensures an accurate representation of road geometries, lane configurations, and traffic signal placements, closely aligning with real-world conditions. The finalized environment is integrated into CARLA and SUMO, establishing a unified platform for vehicle interaction and traffic simulation. While the current simulation models individual vehicle behaviors using CARLA’s vehicle dynamics framework, future work will incorporate SUMO co-simulation to enhance macroscopic and microscopic traffic modeling. This integration will support car-following models, lane-changing behaviors, and signal control strategies, enabling a more comprehensive representation of large-scale traffic dynamics.

Fig.~\ref{fig_digitaltwin}(c) presents the DT environment, showcasing road markings, buildings, and trees, contributing to an immersive and realistic simulation. This high-fidelity rendering ensures that the DT closely resembles real-world conditions. Meanwhile, Fig.~\ref{fig_digitaltwin}(d) highlights an intersection as an example, where an RSU twin is mounted on a traffic signal pole. The green line in the figure represents active V2X communication, illustrating BSM transmissions from the OBU twin and SPaT/MAP messages from the RSU twin. These interactions ensure that the OBU twin can formulate the current signal phase state, including signal group details, movement states, and remaining time.

\begin{figure*}[!ht]
  \centering
  \includegraphics[width=0.8\textwidth]{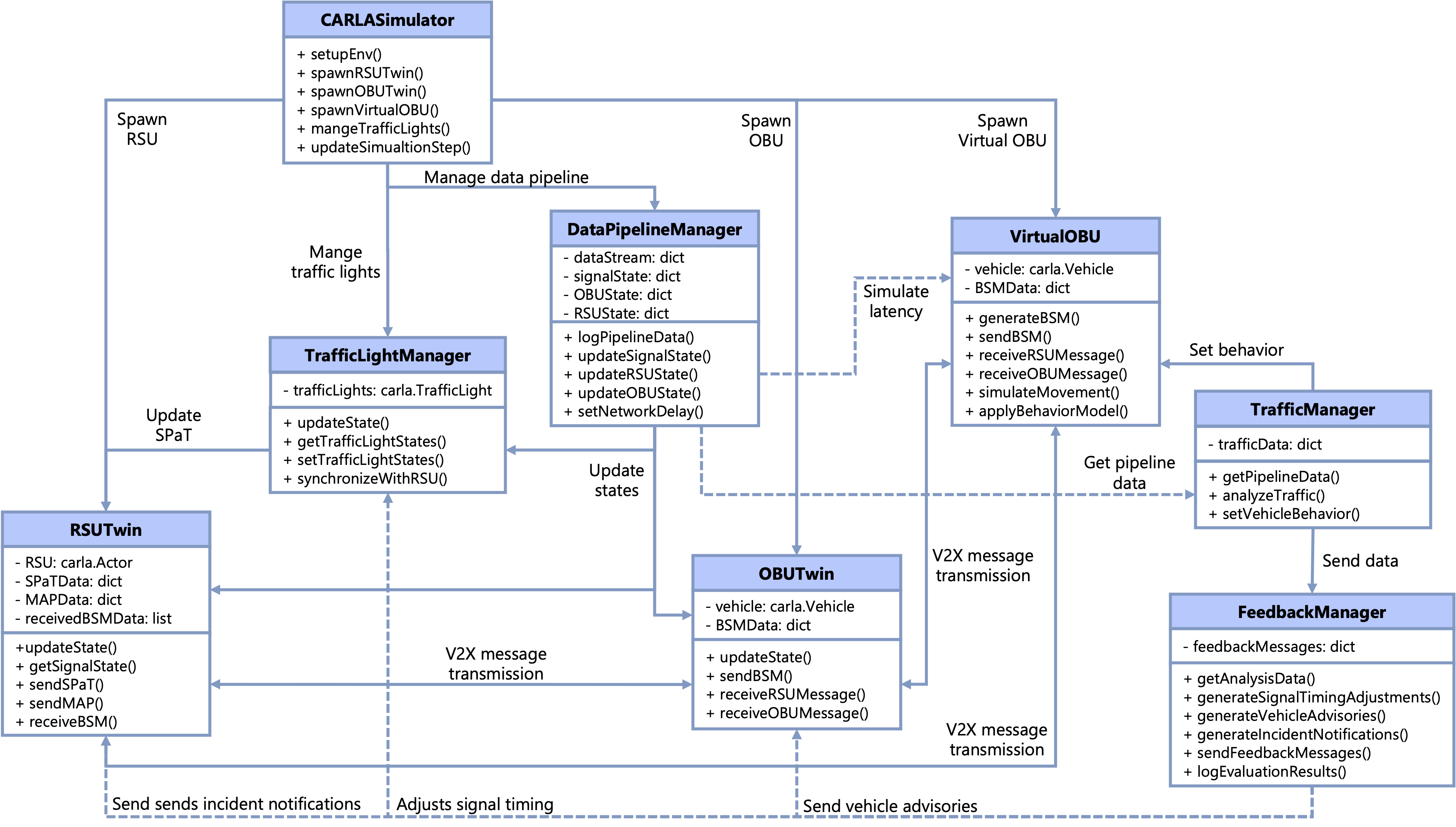}
  \caption{V2X Communication UML Class Diagram in CARLA}\label{fig_v2x_communication_uml}
\end{figure*}


\subsubsection{V2X System Design in CARLA}
To simulate V2X communication, we designed a framework that models information exchanges between OBUs, RSUs, and signal controllers. Fig.~\ref{fig_v2x_communication_uml} presents the UML class diagram representing the modular structure of V2X interactions in the DT~\cite{grimm2024co}. The V2X communication system consists of distinct modules. The \texttt{DataPipelineManager} is responsible for ingesting real-world V2X messages and updating the states of RSUs, OBUs, and traffic signals. The \texttt{RSUTwin} processes and transmits SPaT and MAP messages, ensuring that vehicles receive accurate signal phase and intersection geometry information. Meanwhile, the \texttt{OBUTwin} generates BSM messages based on the real-time state of the vehicle, allowing the system to simulate and analyze vehicle interactions. The \texttt{VirtualOBU} enhances the DT by generating synthetic BSM messages, which are then transmitted to RSUs and other OBUs to simulate a realistic environment. The \texttt{TrafficLightManager} regulates traffic light states and ensures synchronization with the corresponding \texttt{RSUTwin} at the intersection to maintain consistency in signal operations.

In the DT, V2X message transmission occurs through structured interactions between RSUs and OBUs. The \texttt{RSUTwin} continuously transmits SPaT and MAP messages to vehicles within its range using CARLA’s API, synchronizing real-world traffic light phases with the simulation. Simultaneously, the \texttt{OBUTwin} updates its state dynamically based on received V2X messages, allowing vehicles to react in real time to changing traffic conditions. The inclusion of a \texttt{NetworkDelay} module introduces latency modeling, ensuring that the simulation accounts for realistic message propagation delays, an essential factor in evaluating the performance of connected vehicle applications. To integrate this feature within the \texttt{DataPipelineManager}, the \texttt{setNetworkDelay()} function is implemented to control latency, packet loss, and communication reliability of V2X messages, ensuring that RSU and OBU interactions accurately reflect real-world transmission conditions. For \texttt{VirtualOBU}, empirical results from field tests—specifically key performance indicators (KPIs) such as latency, packet error rate (PER), and spatial delivery rate—are integrated into the simulation to accurately model communication behavior~\cite{li2023does}.

The \texttt{TrafficManager} retrieves pipeline data, monitors the states of \texttt{RSUTwin}, \texttt{OBUTwin}, and \texttt{VirtualOBU}, and optimizes vehicle behavior based on real-time traffic conditions. Instead of generating control messages, it focuses on analyzing traffic patterns and adapting vehicle interactions accordingly. The \texttt{FeedbackManager} processes traffic insights from \texttt{TrafficManager} and generates control feedback. It manages signal timing adjustments, vehicle advisories, and incident notifications, ensuring effective communication between infrastructure and vehicles. The \texttt{TrafficLightManager} receives signal adjustments, \texttt{OBUTwin} processes advisories, and \texttt{RSUTwin} handles incident alerts, enabling dynamic traffic optimization in the simulation environment.

\subsubsection{Feedback Generation and Integration}

To ensure a bidirectional connection between the virtual and physical environments, the DT generates feedback based on real-time V2X data and transmits actionable insights back to the physical system. These feedback mechanisms enhance the interaction between the digital and physical systems, supporting real-time decision-making for improved safety and mobility. As Fig.~\ref{fig_digitaltwin} shows, the three primary types of feedback include:

\begin{enumerate}
    \item \textbf{Traffic Signal Timing Adjustments}: Based on CV trajectory data, queue length estimations, and SPaT messages, the virtual system generates recommendations to optimize signal phases. If a CV consistently encounters prolonged stops at an intersection, the system analyzes queue lengths by measuring the distance between the stopped vehicle and the stop line. When excessive delays are detected, localized green-phase extensions are suggested to improve signal coordination for that approach.

    \item \textbf{Vehicle Advisory Messages}: The DT dynamically computes advisory speed recommendations based on the vehicle’s distance to the stop line and the remaining green time. If a CV is approaching an intersection with an imminent phase change, the system calculates an optimal speed to help the vehicle either clear the intersection safely or smoothly decelerate.

    \item \textbf{Incident Notifications}: The system continuously monitors CV motion states and detects unsafe driving behaviors, such as erratic speed fluctuations, sudden braking, and deviations from expected lane trajectories. If a CV exhibits abnormal behavior, such as a rapid deceleration from high speed or an unanticipated trajectory change, the system flags a potential hazard.
\end{enumerate}

The feedback mechanism operates through the RSU’s web application, broadcasting feedback messages to OBUs and traffic signal controllers. By integrating traffic signal optimizations, vehicle advisories, and hazard alerts, the DT actively influences physical infrastructure, ensuring a continuous feedback loop between simulation and real-world deployment.

\section{Experiment Visualization and Simulation}
We have conducted multiple experiments between April and July 2023. Six RSUs and a Cohda MK6C OBU were used for collecting data. The data pipeline was employed to retrieve and decode V2X messages from RSUs. Data was collected from the OBU using a laptop via an Ethernet connection, enabling real-time processing.

\subsection{Data Integration and Visualization}
\begin{figure*}[!ht]
  \centering
  \includegraphics[scale=0.5]{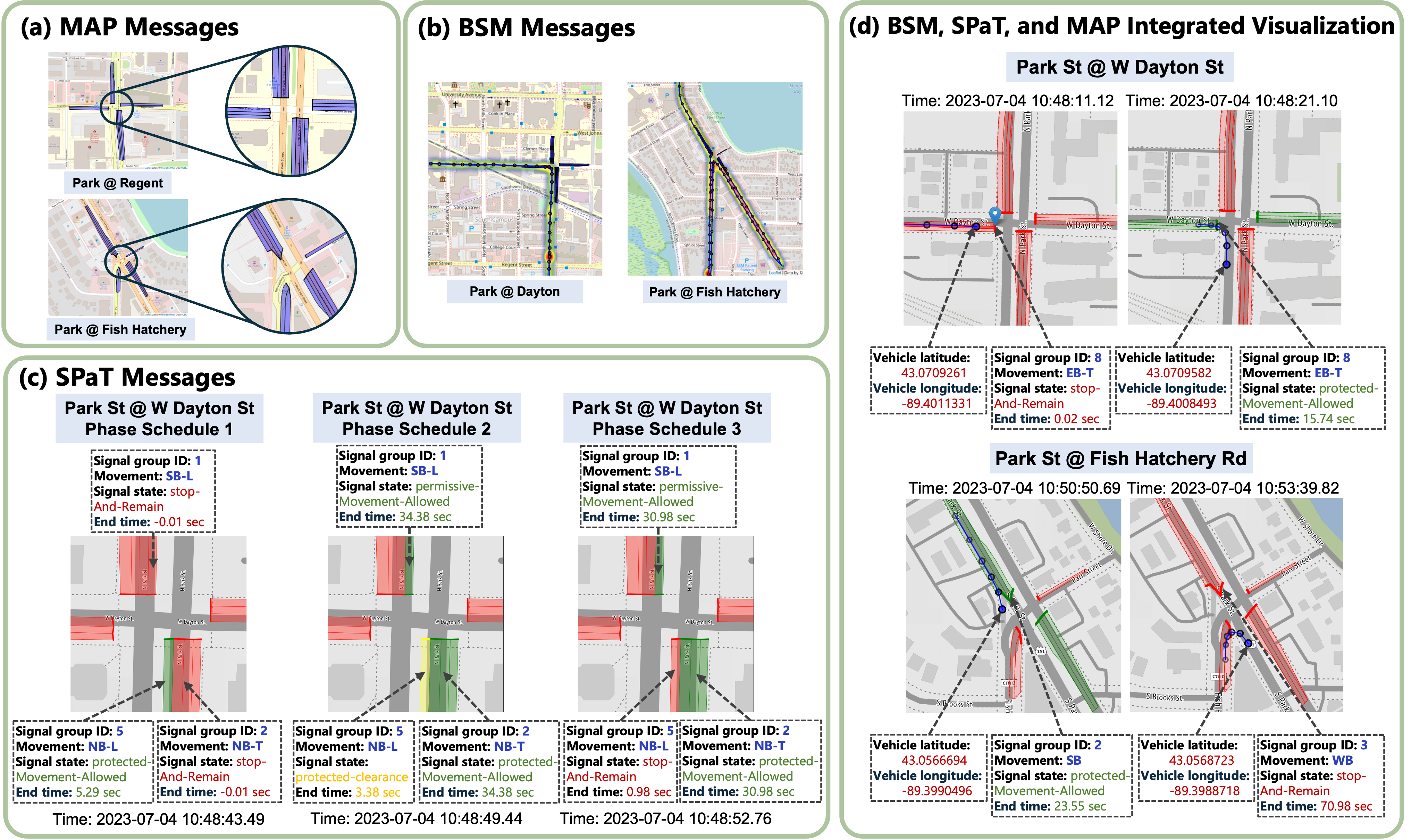}
  \caption{V2X Data Integration Visualization. (a) Visualization of MAP Messages. (b) Visualization of BSM Messages. (c) Visualization of SPaT Messages.  (d) BSM, SPaT, and MAP Integrated Visualization. }\label{fig_data_viz}
\end{figure*}

Fig.~\ref{fig_data_viz} (a) shows the CV corridor's geographic layout. Insets visualize MAP messages at two sample intersections, showing detailed lane configurations and traffic signal positions. Moreover, Fig.~\ref{fig_data_viz} (b) visualizes the trajectory of a vehicle using BSM messages. The vehicle’s trajectory is marked by blue lines, the surrounding polygons depict the lane geometries, and the heatmap overlay provides a visual representation of the vehicle’s speed, with red areas indicating slower speeds and green areas marking faster movement. In addition, we visualize SPaT and MAP messages to understand signal phases and intersection geometries. As shown in Fig. \ref{fig_data_viz} (c), a three-phase signal schedule is applied, with red, yellow, and green colors representing the status of traffic signals. For instance, the red represents stop and the green stands for permissive or protected movement. Integrating SPaT and MAP messages is crucial for assessing how well the vehicles align with traffic signals and lane configurations. This information can help traffic engineers optimize phase schedules to reduce congestion, particularly at complex intersections.

Lastly, Fig.~\ref{fig_data_viz} (d) integrates BSM, SPaT, and MAP messages, presenting a comprehensive view of the physical system. This figure highlights the real-time position of the vehicle (denoted by blue points) relative to the lane geometries. The lane currently occupied by the vehicle is marked with a distinct color, reflecting the traffic signal phase. The integration of real-time V2X messages offers integrated information about infrastructure, vehicles, and signals, providing a solid foundation for developing the DT.

\subsection{Digital Twin Simulation Experiments}

\begin{figure}[!ht]
  \centering
  \includegraphics[width=0.48\textwidth]{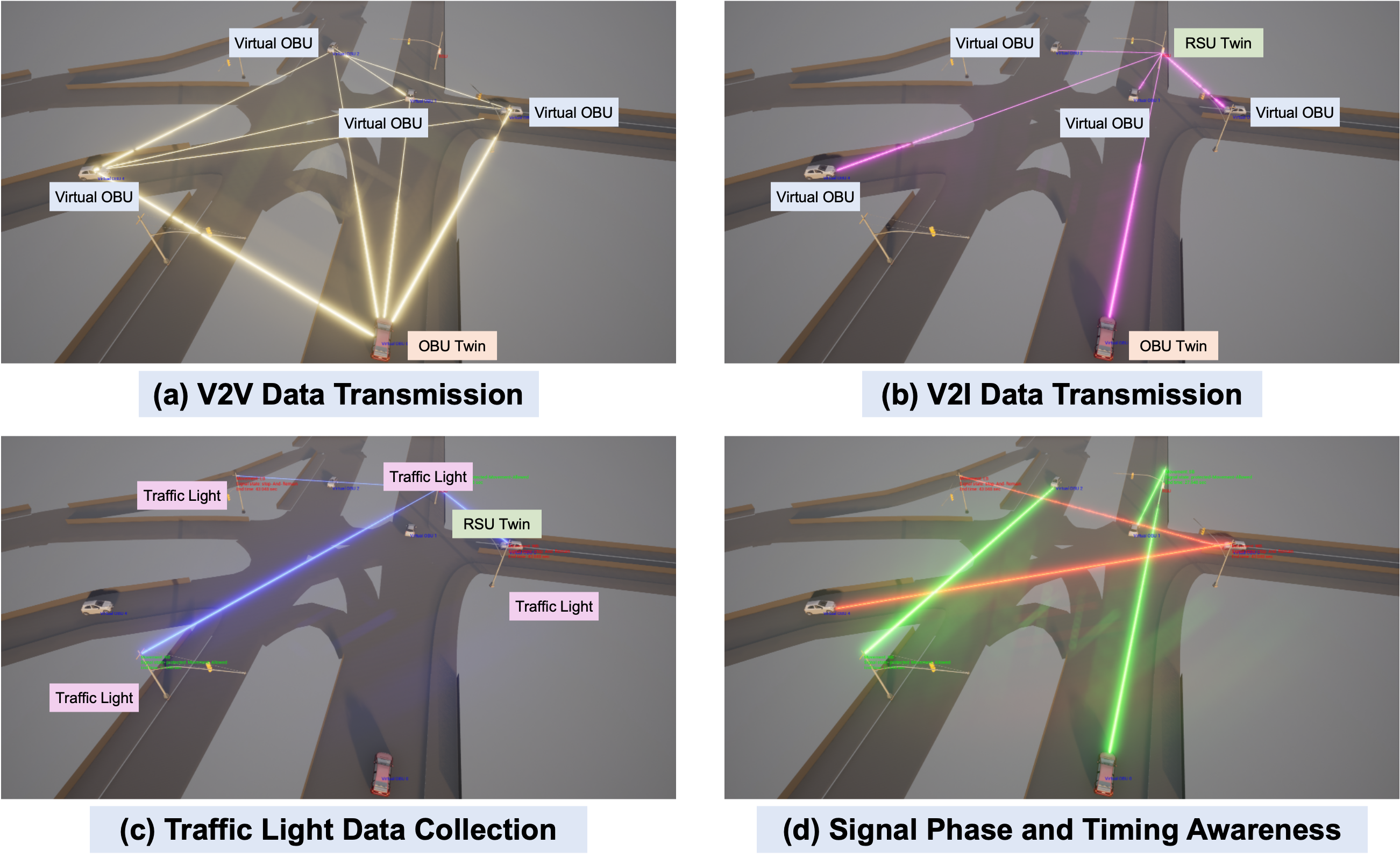}
  \caption{V2X Data Transmission Simulation in the DT. (a) V2V Communication: OBU Twin and Virtual OBUs. (b) V2I Communication: OBUs and RSU Twin. (c) Traffic Signal Data Collection by RSU Twin. (d) Signal Phase and Timing Awareness.}\label{fig_v2x_sim}
\end{figure}

Fig.~\ref{fig_v2x_sim} illustrates V2X data transmission and interaction within the DT. In Fig.~\ref{fig_v2x_sim}(a), V2V communication occurs between the OBU Twin and multiple Virtual OBUs, enabling cooperative maneuvers and trajectory adjustments. Fig.~\ref{fig_v2x_sim}(b) shows V2I communication, where OBUs exchange data with the RSU Twin, receiving SPaT messages and transmitting BSMs for real-time traffic updates. Fig.~\ref{fig_v2x_sim}(c) depicts traffic signal data collection, where the RSU Twin retrieves real-time SPaT messages from traffic controllers, ensuring vehicles receive accurate signal phase updates. In Fig.~\ref{fig_v2x_sim}(d), the green lines represent the active signal phases that vehicles recognize and respond to.

\begin{figure*}[!t]
    \centering
    \begin{minipage}{0.48\textwidth}
        \centering
        \includegraphics[width=\textwidth]{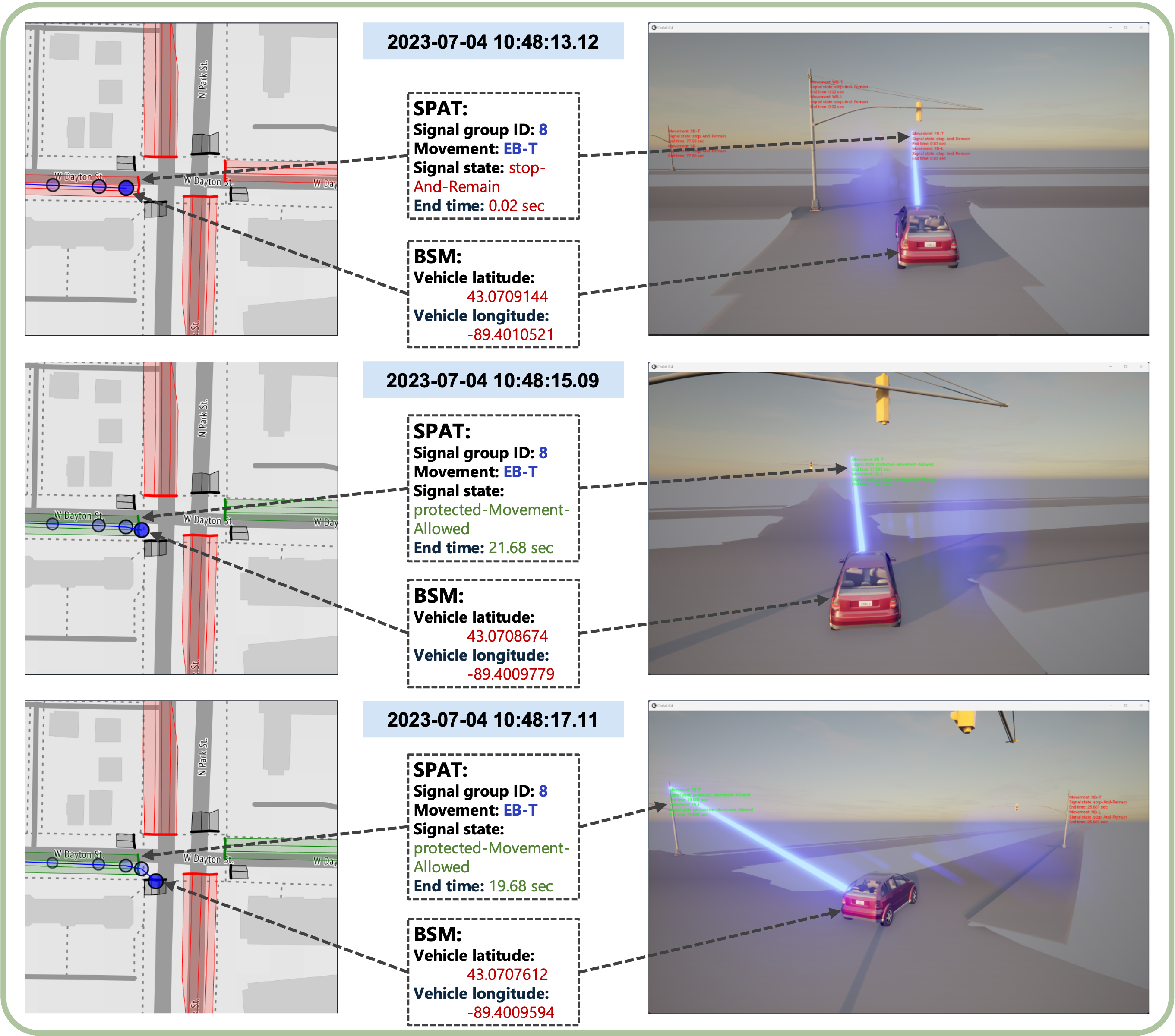}
    \end{minipage}
    \hfill
    \begin{minipage}{0.48\textwidth}
        \centering
        \includegraphics[width=\textwidth]{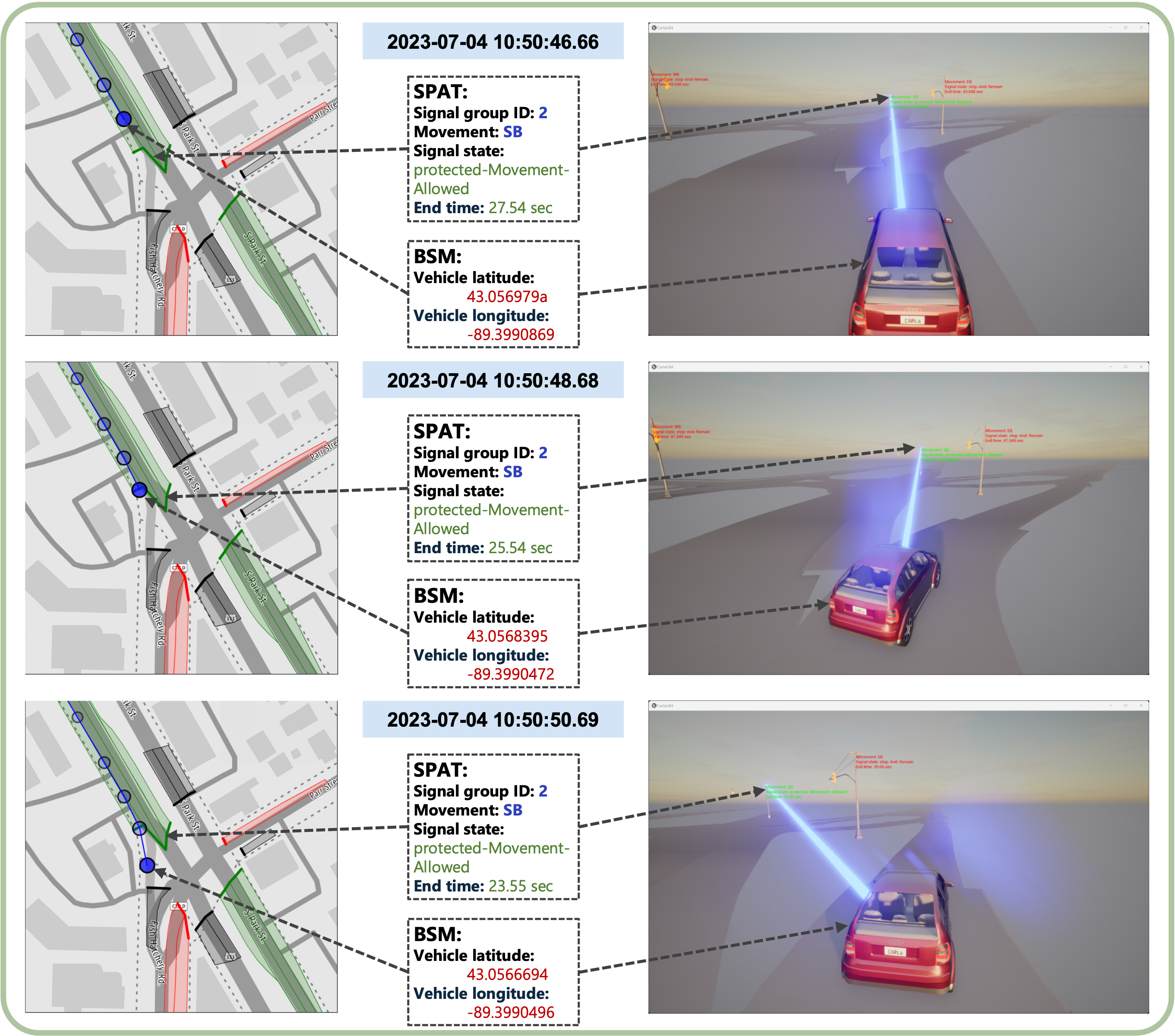}
    \end{minipage}
    \caption{Simulation Performance Evaluation Experiments: Park @ Dayton Intersection (left), and Park @ Fish Hatchery Intersection (right)}\label{fig_dt_exp}
\end{figure*}

Moreover, we conducted additional experiments to evaluate the DT's fidelity in replicating real-world traffic conditions. The first experiment in Fig.~\ref{fig_dt_exp} was conducted at the intersection of Park Street and West Dayton Street. This figure presents a series of snapshots at different timestamps, demonstrating how the DT replicates vehicle movement and signal interaction. The left figure shows the real-world intersection view, highlighting vehicle positions and the corresponding signal phases. On the right side, the DT reflects the real-time behavior of a virtual vehicle and RSU, synchronizing the vehicle behavior and traffic light states with the physical world. For instance, at timestamp 2023-07-04 10:48:13.12, the vehicle stops at a red light (EB-T), as indicated by the red signal in both the physical and virtual environments. The corresponding SPaT message confirms a stop-and-remain command with a remaining time of 0.02 seconds, showing that the signal is about to change. In subsequent snapshots at 10:48:15.09 and 10:48:17.11, the light transitions to green (protected movement), allowing the vehicle to move forward. The vehicle's position and trajectory closely align between the real-world data and the simulation, demonstrating accurate synchronization and replication of movement and signal phases.

The second experiment took place at the intersection of Park Street and Fish Hatchery Road. The three timestamps capture a vehicle’s movement in the southbound approach (SB) during a protected phase with a green light. Positioned in the southbound right-turn lane, which shares a signal group with the southbound through lane, the DT tracks the vehicle as it progresses through the intersection while the through signal remains in the protected phase. Additionally, SPaT messages confirm that the signal phase permits the vehicle to proceed, starting with 27.54 seconds remaining at the first timestamp and gradually counting down as the vehicle moves forward.

In summary, the DT accurately replicates vehicle behavior, signal phases, and timings in both experiments, demonstrating its high fidelity. The DT enables precise emulation of vehicle movements and infrastructure interactions, which is critical for T-CPS. 

\subsection{Feedback Analysis}
\begin{figure}[!ht]
    \centering
    \includegraphics[width=0.3\textwidth]{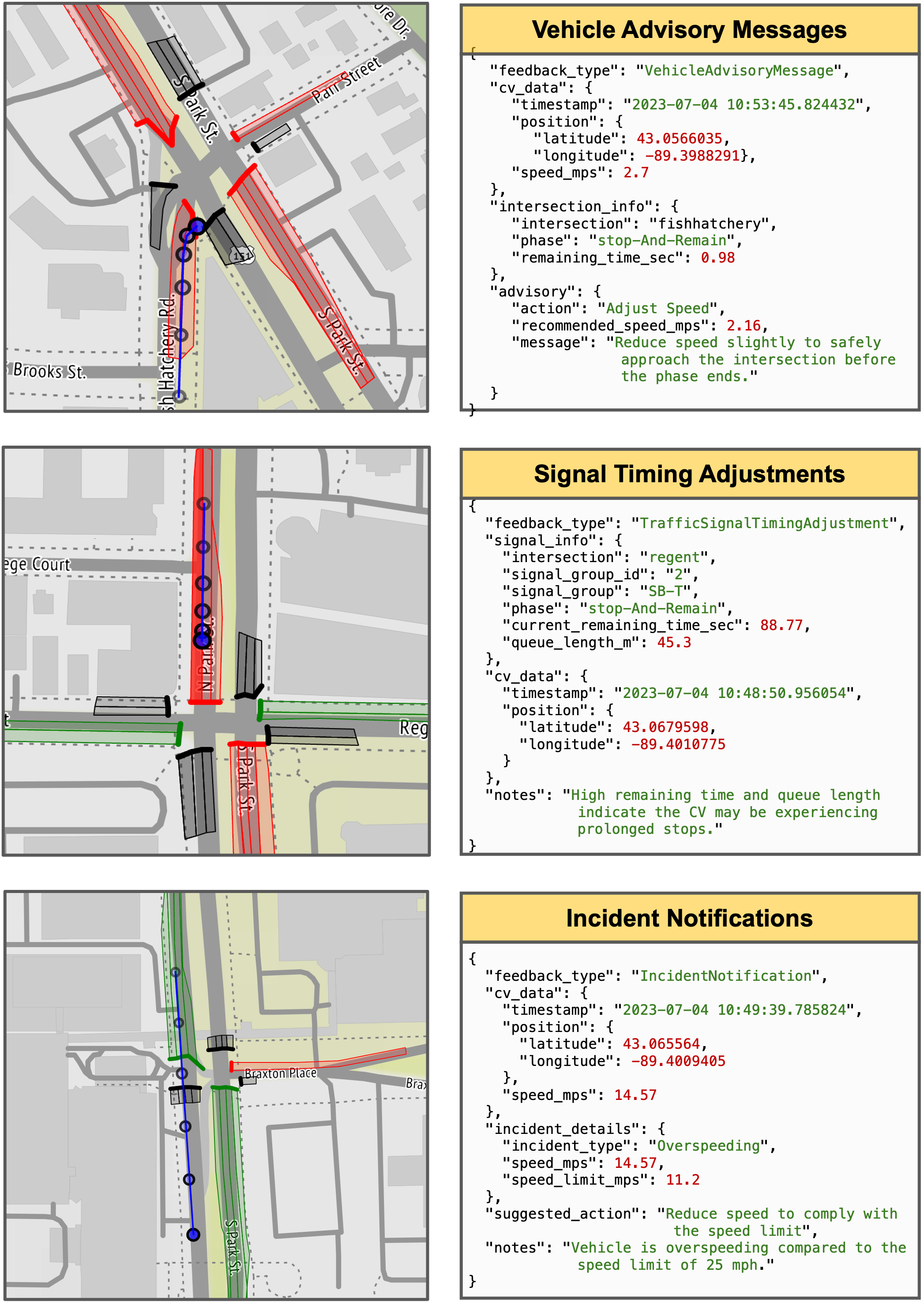}
    \caption{Feedback Analysis Results}
    \label{fig_feedback}
\end{figure}

Fig.~\ref{fig_feedback} illustrates the real-time feedback generated by the DT, which continuously monitors vehicle movements and intersection conditions through V2X data. Each feedback type is visualized with corresponding traffic conditions and structured JSON outputs, demonstrating the system's capability to enhance traffic safety and efficiency.

In the first scenario, when a vehicle approaches the intersection Park @ Fish Hatchery at 2.7 m/s, with less than one second remaining before the signal phase transition, the system evaluates the situation and recommends a speed adjustment to 2.16 m/s. This adjustment ensures the vehicle can safely approach the intersection without abrupt braking or unnecessary acceleration. At the intersection Park @ Regent controlled by the southbound-through (SB-T) signal group, the system detects a queue length of 45.3 meters and a remaining red time of 88.77 seconds, indicating prolonged vehicle stoppage. In response, a signal timing adjustment is flagged to optimize traffic flow and mitigate excessive delays. In the third scenario, a vehicle traveling at 14.57 m/s (32.6 mph) in a speed-restricted zone of 11.2 m/s (25 mph) triggers an overspeeding notification. The system detects the violation and issues an advisory instructing the vehicle to reduce speed to comply with the limit. 

These results demonstrate the DT’s capability to dynamically generate actionable feedback, supporting adaptive traffic control and real-time safety interventions. By integrating vehicle-specific advisories, signal phase optimization, and automated incident detection, the system contributes to safer, more efficient traffic operations.

\section{Discussions} 

\subsection{System Highlights} 
\subsubsection{\textbf{System Scalability}}
Scalability is a top priority as the CV corridor expands with additional RSUs. Leveraging distributed edge computing allows edge servers to process data locally, maintaining low latency and improving overall system performance. Additional servers can be configured as needed to manage new RSUs, reducing single points of failure and improving the system’s resilience. This distributed architecture not only ensures scalability but also simplifies error diagnosis and recovery, enhancing the system’s robustness.

\subsubsection{\textbf{System Redundancy}}
The DT incorporates redundancy to ensure continuous operation. In case of an edge server failure, the system can quickly reroute tasks to a backup server, minimizing downtime. Moreover, overlapping RSU coverage allows for multiple RSUs to collect the same V2X messages, enhancing data reliability. This redundancy is crucial for maintaining uninterrupted communication and ensuring consistent data availability for real-time decision-making in traffic management.

\subsubsection{\textbf{Data Integrity}}
The multi-point connection between OBUs and RSUs enhances data integrity by allowing multiple RSUs to receive the same messages, ensuring seamless data transitions as OBUs move between zones. Cross-validation techniques, like majority voting, ensure data accuracy even in cases of communication discrepancies. This approach guarantees that the DT provides a reliable, real-time representation of the CV corridor, which is critical for accurate traffic management and analysis.


\subsection{Applications}
\subsubsection{\textbf{Behavior monitoring}} 
The DT precisely determines each vehicle’s lane position, enhancing the granularity of traffic monitoring. This level of spatial awareness allows traffic managers to monitor the behaviors of individual vehicles and identify dangerous behaviors such as red-light violations and wrong-way driving.
\subsubsection{\textbf{Signal broadcasting and optimization}}
The DT offers real-time updates on traffic signal status. This information can be transmitted to human-driven vehicles, AVs, and CVs for timely, informed decision-making, optimizing traffic flow by allowing vehicles to adjust their speeds and reducing the need for stops at intersections. Additionally, the integration of real-time vehicle-level data enhances signal optimization, further improving traffic flow and minimizing congestion.
\subsubsection{\textbf{Traffic planning}} The DT archives long-term traffic condition data, enabling traffic managers to analyze historical patterns and trends. This capability supports the development of strategic planning initiatives aimed at improving future traffic conditions, optimizing roadway design, and informing infrastructure investments.

\subsection{Limitations}
\subsubsection{\textbf{Simulated Traffic and Virtual OBUs}}
Due to the absence of additional sensors on CVs, capturing detailed surrounding traffic conditions and segment-level flow is challenging in real-world deployments. To compensate, simulated traffic and virtual OBUs are introduced in the DT. However, simulated vehicles follow predefined models, which may not fully replicate real driver behavior, unexpected traffic variations, or sensor uncertainties.
\subsubsection{\textbf{Communication Delay Modeling}}
In a real-world environment, V2X message transmission is subject to network congestion, packet loss, and variable latency. In the DT, communication delays between RSU twins and OBU twins can be closely mapped using real-world data, reflecting actual message transmission and reception gaps. However, for virtual OBUs, communication delay must be statistically simulated based on field experiments, which may not capture all network fluctuations.
\subsubsection{\textbf{Perception Constraints and Simulated Sensing}}  
The lack of onboard perception sensors on CVs and RSUs limits real-world detection of road users and environmental conditions. In contrast, the CARLA simulation environment allows virtual CVs to be equipped with cameras, LiDAR, and radar, enabling perception-based V2V and V2I communication. This extends forecasting capabilities to dynamic obstacle detection, lane occupancy prediction, and cooperative perception.
\subsubsection{\textbf{Real-World Validation and Hybrid Testing}}
To mitigate discrepancies between simulated and real-world environments, continuous validation with real-world data is necessary. Hybrid testing, where real OBUs interact with simulated traffic, is an area of future research to enhance the realism and reliability of DT-generated insights.

\subsection{Future Improvements}
\subsubsection{\textbf{Traffic Analysis and Estimation}} By integrating traffic data from WisTransportal \cite{parker2006wistransportal} with real-time V2X data, we can bridge the gap between macroscopic data (e.g., speed, volume, and occupancy) and microscopic vehicle data. This integration enables a deeper understanding of the impact of CVs and CAVs on traffic in mixed traffic flow. For instance, real-time traffic density estimation at intersections, based on BSM and MAP messages, allows for dynamic signal timing adjustments, improving traffic flow and reducing congestion. Furthermore, CVs can receive updates on congestion and alternative routes, further easing traffic. Lastly, the integration aids long-term infrastructure planning, such as lane expansions or new traffic control measures in high-traffic areas.

\subsubsection{\textbf{Integration of Mathematical Traffic Models}} To enhance the fidelity of the DT, future work will incorporate SUMO co-simulation to introduce mathematical models of traffic flow. By integrating car-following models, lane-changing behaviors, and adaptive signal control strategies, SUMO will enable large-scale traffic network modeling. This integration will complement CARLA’s high-fidelity vehicle physics, providing a multi-scale DT that captures both individual vehicle dynamics and system-wide traffic behavior.

\subsubsection{\textbf{Preparation for Multi-OBU Cooperation}} Data sharing between vehicles and infrastructure is enhanced by leveraging the C-V2X communication modules. Testing the communication between RSUs and OBUs equipped with C-V2X modules is a priority, and future experiments will involve deploying additional CVs for this purpose. These experiments will evaluate the cooperative interaction between RSUs and OBUs, setting the stage for more accurate simulations that are essential for the further development of the DT.


\section{Conclusions}
This research presents a DT framework based on a real-world CV corridor, designed as an essential component of T-CPS. By leveraging real-time V2X data, as well as communication, computing, and simulation technologies, the DT achieves high-fidelity replication of physical traffic conditions including vehicle behaviors, signal timing, and lane geometry. In addition, the DT enhances safety and mobility by generating real-time feedback to the physical system, demonstrated through applications such as signal timing adjustments, vehicle advisory messages, and incident notifications. Furthermore, a scalable and redundant data pipeline is integrated to ensure reliability and robustness. The use of distributed edge computing supports real-time data processing while minimizing single points of failure, ensuring the system remains operational under various conditions. The redundant RSU coverage enhances the system’s reliability, allowing continuous data collection and cross-validation, which maintains the integrity and accuracy of the data. The DT sets the stage for expanding CV corridor applications, including large-scale deployment, multi-vehicle interaction modeling, and more refined signal timing optimization.

However, this study also has several limitations that should be addressed in future research. First, the DT is developed for a section of the CV corridor and should be expanded for a comprehensive replication. Second, additional channels, such as smartphones, tablets, infotainment systems, and dashboards, should be explored to communicate information to drivers, traffic managers, and other road users. Moreover, alternative data sources such as cameras, LiDARs, radars, and traffic detectors can be integrated to enhance the accuracy and completeness of the DT.


\section{Acknowledgements}
The Park Street Smart Corridor is being developed through a collaboration of the TOPS Lab, the City of Madison, Traffic and Parking Control Products and Solutions (TAPCO), and the Wisconsin Department of Transportation.  The ideas and views expressed in this paper are strictly those of the Traffic Operations and Safety (TOPS) Laboratory at the University of Wisconsin-Madison. 



\bibliographystyle{IEEEtranN}
\bibliography{ref_new}

\vspace*{-2em}
\begin{IEEEbiography}[{\includegraphics[width=1in,height=1.25in,clip,keepaspectratio]{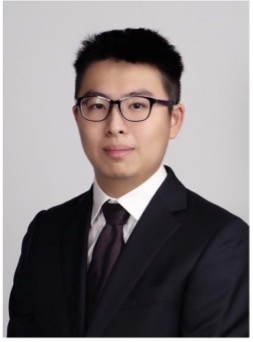}}]{Keshu Wu}
Keshu Wu is currently a postdoctoral research associate at Texas A\&M University. He receives his Ph.D. in Civil and Environmental Engineering from the University of Wisconsin-Madison in 2024. He also holds an M.S. degree in Civil and Environmental Engineering from Carnegie Mellon University in 2018 and an M.S. degree in Computer Sciences from the University of Wisconsin-Madison in 2022. He completed his B.S. in Civil Engineering at Southeast University in Nanjing, China in 2017. His research interests include the application and innovation of artificial intelligence and deep learning techniques in connected automated driving, intelligent transportation systems, and digital twin modeling and simulation.
\end{IEEEbiography}

\vspace*{-2em}
\begin{IEEEbiography}[{\includegraphics[width=1in,height=1.25in,clip,keepaspectratio]{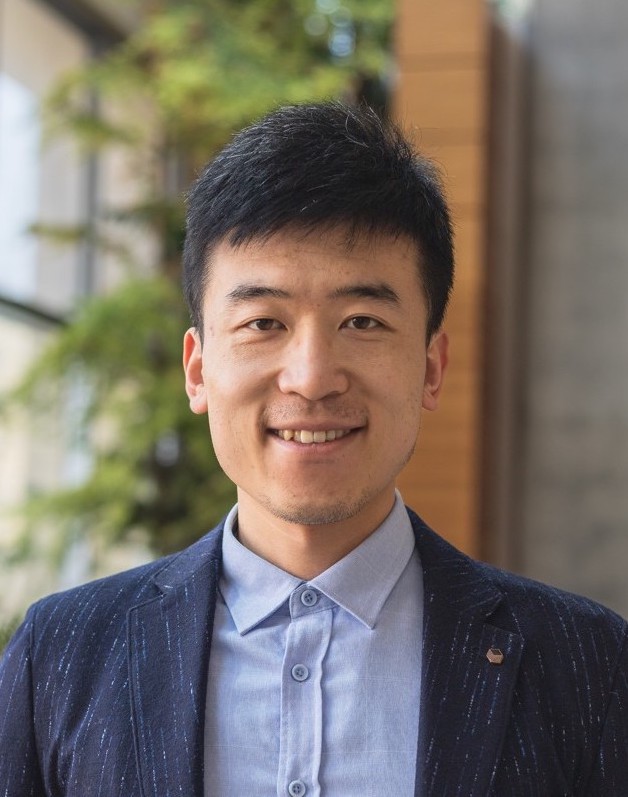}}]{Pei Li}
Dr. Pei Li is a Scientist in the Department of Civil and Environmental Engineering at the University of Wisconsin-Madison. He received his Ph.D. in Civil Engineering with a focus on Transportation Engineering from the University of Central Florida in 2021, after which he served as a Postdoctoral Research Fellow at the University of Michigan Transportation Research Institute. His research interests include transport safety, smart mobility, human factors, machine learning, connected and automated vehicles, and digital twins.
\end{IEEEbiography}

\vspace*{-2em}
\begin{IEEEbiography}[{\includegraphics[width=1in,height=1.25in,clip,keepaspectratio]{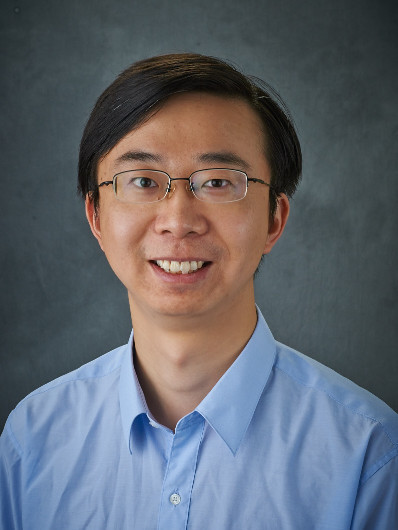}}]{Yang Cheng}
received the B.S. and M.S. degrees in automation from Tsinghua University, Beijing, China, in 2004 and 2006, respectively, and the Ph.D. degree in civil engineering from the University of Wisconsin–Madison in 2011. He is currently a scientist at the Wisconsin Traffic Operations and Safety (TOPS) Laboratory of the University of Wisconsin-Madison. His research areas include automated highway and driving systems, mobile traffic sensor modeling, large-scale transportation data management and analytics, and traffic operations and control.
\end{IEEEbiography}

\vspace*{-2em}
\begin{IEEEbiography}[{\includegraphics[width=1in,height=1.25in,clip,keepaspectratio]{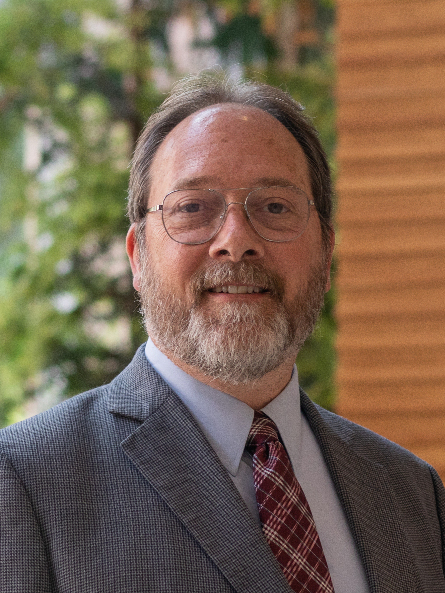}}]{Steven T. Parker}
is the Managing Director of the Wisconsin Traffic Operations and Safety (TOPS) Laboratory at the University of Wisconsin-Madison. He has led a range of research and development initiatives for the TOPS Lab across several core areas including transportation safety, work zone systems, traffic management systems, and connected and automated vehicle technologies. He received a Ph.D. in Computer Science from the University of Wisconsin-Madison. He is currently serving in his second term as the Chair of the Transportation Research Board (TRB) AED30 Information Systems and Technology Committee.
\end{IEEEbiography}

\vspace*{-2em}
\begin{IEEEbiography}[{\includegraphics[width=1in,height=1.25in,clip,keepaspectratio]{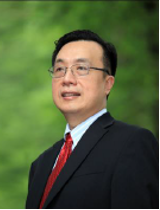}}]{Bin Ran}
is the Vilas Distinguished Achievement Professor and Director of ITS Program at the University of Wisconsin at Madison. Dr. Ran is an expert in dynamic transportation network models, traffic simulation and control, traffic information system, Internet of Mobility, Connected Automated Vehicle Highway (CAVH) System. He has led the development and deployment of various traffic information systems and the demonstration of CAVH systems. Dr. Ran is the author of two leading textbooks on dynamic traffic networks. He has co-authored more than 240 journal papers and more than 260 referenced papers at national and international conferences. He holds more than 20 patents of CAVH in the US and other countries. He is an associate editor of Journal of Intelligent Transportation Systems.
\end{IEEEbiography}

\vspace*{-2em}
\begin{IEEEbiography}[{\includegraphics[width=1in,height=1.25in,clip,keepaspectratio]{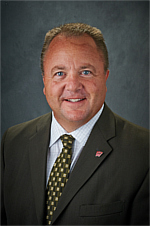}}]{David A. Noyce}
received his B.S. and M.S. degrees in Civil and Environmental Engineering from UW-Madison in 1984 and 1995, respectively, and received his Ph.D. degree in Civil (Transportation) Engineering from Texas A\&M University in 1999. He has authored more than 380 refereed scholarly papers, conference proceedings, research reports, and book chapters. He was elected Fellow in the American Society of Civil Engineers (ASCE) in 2017 and was President of ASCE’s Transportation and Development Institute (T\&DI) in 2022. He works with the National Academy of Sciences and the Transportation Research Board (TRB), where he has chaired several National Cooperative Highway Research Program (NCHRP) project panels and has (and is currently) conducted NCHRP research.
\end{IEEEbiography}

\vspace*{-2em}
\begin{IEEEbiography}[{\includegraphics[width=1in,height=1.25in,clip,keepaspectratio]{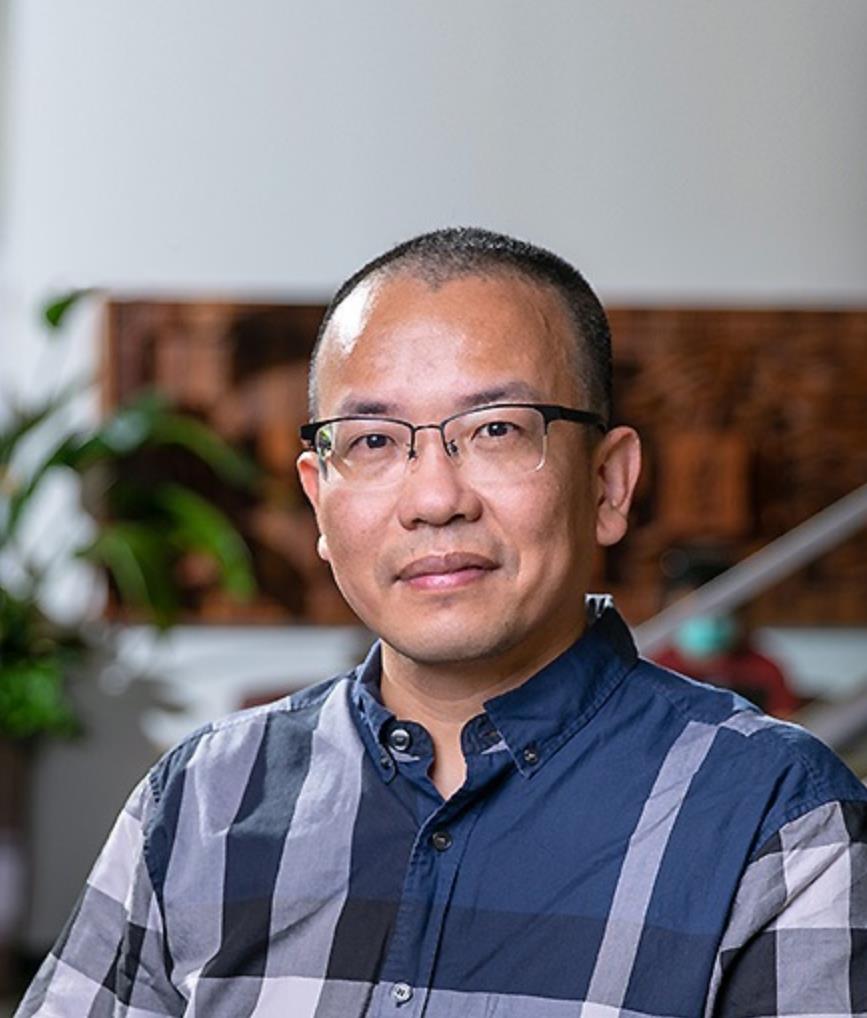}}]{Xinyue Ye}is the Harold Adams Endowed Professor at Texas A\&M University (TAMU) and serves as Faculty Fellow of Strategic Initiatives for The Division of Research. His research, supported by 13 federal agencies and numerous industry partners, integrates computational social science, urban data science, and geospatial AI to address issues like infrastructure resilience, climate change, and social justice. As the first elected Fellow of the American Association of Geographers in the Early/Mid-Career category, he directs the Center for Geospatial Sciences, Applications, and Technology at TAMU.

\end{IEEEbiography}

\end{document}